\documentclass[11pt]{article}

\usepackage[utf8]{inputenc}
\usepackage[T1]{fontenc}
\usepackage{lmodern}
\usepackage[margin=1in]{geometry}
\usepackage{graphicx}
\graphicspath{{figures/}}
\usepackage{booktabs}
\usepackage{amsmath}
\usepackage{amssymb}
\usepackage{xcolor}
\usepackage{microtype}
\usepackage{natbib}
\usepackage[colorlinks=true,allcolors=blue!55!black]{hyperref}
\usepackage{url}

\usepackage{etoolbox}
\AtBeginEnvironment{tabular}{\footnotesize}
\title{\bfseries Progressive Disclosure for LLM-Maintained Wiki\\ Knowledge Bases: a Preregistered Ablation}
\author{Theodore O. Cochran\\ \small AI for Altruism (A4A)\\ \small \href{mailto:theo@ai4altruism.org}{theo@ai4altruism.org}}
\date{}

\newcommand{\corpusnote}{ (\texttt{offload})}
\newcommand{\osflink}{OSF \url{https://osf.io/feka7} (DOI 10.17605/OSF.IO/FEKA7)}
\newcommand{\materialslink}{\url{https://osf.io/a53mz}}

\newcommand{\bibpath}{refs}

\begin{document}
\maketitle

\begin{abstract}
LLM agents now often answer questions from knowledge bases they help maintain. A common intuition says progressive disclosure should make this cheaper. Instead of loading one large index, the agent reads a compact catalog and one-line page summaries, then opens only the pages it needs. We tested that intuition in a preregistered study on a real 709-page markdown knowledge base maintained by an LLM. We retrofitted it for progressive disclosure and built four versions that differ only in how the agent reaches the pages. The pages themselves are identical in every version, so any difference comes from the access structure alone. Each version was tested three ways, with the agent following a set protocol, choosing its own path, or made to load the catalog first. A judge from a different model family graded the answers blind against verified reference answers.

A preparatory pilot changed the question. A capable agent never loaded the large index at all. It worked out from the question where a page was and read it directly. The saving we set out to measure did not exist for such an agent, so we made answer quality the primary outcome. Quality held. Answers from the retrofitted knowledge base were as good as answers from the original, within a margin we set in advance. Two limits apply. Our human rater and the model judge agreed far less than the plan required, so the quality result rests on the judge, backed by sensitivity checks. Quality was also not shown to hold when the agent was forced to load the catalog first, or on the two most reliably graded criteria under a stricter test. Cost fell clearly in every condition we tested, and the retrofitted version cited fewer pages and took fewer tool turns per answer.
\end{abstract}

\section{Introduction}
\label{sec:intro}

Retrieval over a knowledge base is a core primitive for LLM agents, and a widespread design instinct is that \emph{less context is cheaper and no worse}: if an agent can be handed a compact catalog and short per-item summaries, it should be able to route to the few relevant items and answer without ingesting a large index or many full documents. This ``progressive disclosure'' intuition, a term borrowed from user-interface design \citep{nielsen2006}, motivates a range of practical choices (table-of-contents prompts, per-document abstracts, and lightweight retrieval tools), and it underlies a common expectation that restructuring a corpus for leaner access yields large token savings. That expectation is what we put to the test.

We subject that intuition to a controlled test on a real artifact: a 709-page markdown wiki knowledge base, its pages cross-linked into a navigable graph, that has been continuously maintained by an LLM. We retrofit the knowledge base for progressive disclosure and ask two questions. First (RQ1), does the restructuring change the \textbf{quality} of an agent's answers relative to a conventional index-catalog baseline? Second (RQ2), does it change the \textbf{cost} of answering, and does that effect depend on how the agent is allowed to access the corpus?

Two features make the study unusually clean. (i)~The corpus is frozen as immutable git tags, and a content-parity gate verifies that page bodies are byte-identical across arms, so any measured difference is attributable to access structure rather than content. (ii)~The design, hypotheses, replicate count, and analysis plan were preregistered before the confirmatory runs, informed by a small preparatory pilot.

\begin{figure}[t]
\centering
\includegraphics[width=\textwidth]{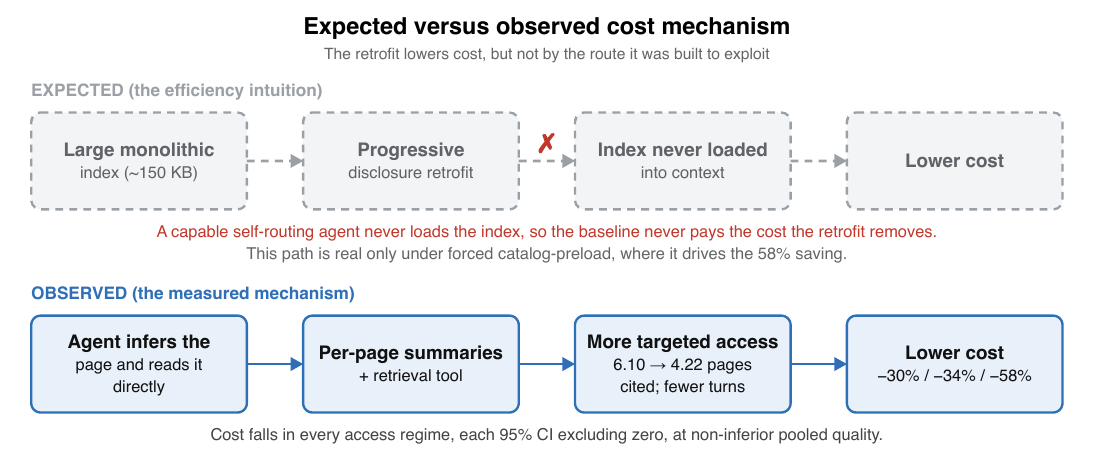}
\caption{Expected versus observed cost mechanism. The efficiency intuition routes the saving through an index that the retrofit makes unnecessary to load; a capable self-routing agent never loads it, so the baseline never pays that cost, and the expected path is real only under forced catalog-preload. The measured saving instead runs through more targeted access. Cost deltas are the A3-vs-A0 contrasts of \S\ref{sec:cost}; pages cited are from \S\ref{sec:perstratum}.}
\label{fig:mechanism}
\end{figure}

The pilot produced the paper's central surprise and reshaped the design. The retrofit was built to eliminate the cost of loading a $\sim$150~KB index on every query, nominally a $\sim$99\% saving on that step. But a capable tool-using agent \textbf{never loads the index}. Our answering model is Claude Opus 4.8, given file-read and search tools and left to choose which to invoke; presented with a question, it infers the likely page path and reads that page directly, so the baseline never pays the cost the retrofit was built to remove. That saving materializes only in a ``catalog-preload'' regime, where the harness concatenates the whole index into the prompt instead of letting the agent fetch on demand. This does not settle whether the retrofit changes total cost. A retrieval tool and per-page summaries might still cut cost by leading the agent to open fewer or smaller pages, and measuring that is exactly what our per-condition cost analysis is for. So we reframed the study before locking it: \textbf{answer quality is the primary outcome}, and we report cost by access condition. The gap between the nominal saving and the measured one, an efficiency claim that does not survive contact with how the system is actually used, is itself a contribution, and it mirrors the evaluation-validity discipline we bring to grading the study. Figure~\ref{fig:mechanism} contrasts the expected mechanism with the one we measure.

\textbf{Contributions:}
\begin{enumerate}
\item A preregistered, content-parity-controlled ablation isolating the causal effect of knowledge-base \emph{access structure} (not content) on LLM answer quality and cost, on a real maintained corpus.
\item A finding that the cost benefit of progressive disclosure is \textbf{real in every access regime} (roughly a third for a capable self-routing agent, 58\% under forced catalog-preload), but arises from a different mechanism than the intuition assumes. It does not come from avoiding a monolithic index load, which a capable self-routing agent never incurs, but is associated with more targeted access behavior: fewer pages cited and fewer tool turns per answer. This contradicts the preregistered cost-null prediction for self-routing agents: A3 reduces cost under enforced and free access as well as under catalog preload.
\item A reusable open harness and audit trail for restricted-corpus evaluation (immutable-tag arms, blinded cross-family grading, hashed held-out gold, aggregate result release) that applies threat-to-validity analysis to its own tooling.
\end{enumerate}

\section{Background and Related Work}

\textbf{Retrieval-augmented generation and context construction.} RAG systems assemble context for an LLM from a corpus \citep{lewis2020,gao2023}, trading off recall against context length and cost \citep{fan2024}. Much of this literature optimizes \emph{what} to retrieve; we instead hold content fixed and vary the \emph{structure} through which an agent accesses it, isolating the structural contribution. Preloading a large catalog also risks the ``lost in the middle'' effect, in which models underuse information buried in long contexts \citep{liu2024}; our forced condition puts that pattern to the test. Recent ``context engineering'' practice, including LLM-maintained knowledge bases \citep{karpathy2026}, emphasizes compact catalogs and summaries; our forced condition operationalizes the naive version of that pattern (preload the catalog) as one of three access regimes. Prior work benchmarked this same LLM-compiled-wiki architecture head-to-head against single-round Vector RAG on a small multi-domain corpus \citep{cochran2026}. There, the wiki scored higher on multi-paper synthesis under LLM judges while spending about 21 times more tokens per query, a decomposition-retrieval variant of RAG recovered most of that synthesis gap at lower cost, and a citation-support advantage for the wiki was not established once the automated scorer was checked against human annotators. That study recommends reporting synthesis structure, citation support, and cost separately, and the recommendation motivates our separation of quality (primary) from cost (secondary) here; whereas \citet{cochran2026} compared two \emph{architectures}, the present study holds the LLM-compiled-wiki architecture fixed and varies only its internal \emph{access structure}. We stress that this forced regime is \emph{not} embedding-similarity RAG: it performs no vector retrieval and no similarity-based selection. It preloads a monolithic catalog (arms A0--A2, the original or slimmed index) or a keyword-ranked result (A3) into the prompt, isolating the \emph{preload-into-context} pattern rather than a particular retrieval algorithm.

\textbf{LLM agents that read their own tools.} Modern coding/agent harnesses give models file-system tools (read, glob, grep) and let them decide what to open. A recurring (and, we argue, under-measured) consequence is that such agents \emph{self-route}: they guess file locations and read targeted content rather than ingesting an index. Our conditions make this explicit by contrasting a protocol-constrained agent, a free self-routing agent, and a forced catalog-preload agent.

\textbf{LLM-as-judge and its threats.} Using an LLM to grade free-text answers is now common \citep{zheng2023}, but LLM judges are subject to self-preference \citep{panickssery2024}, sycophancy \citep{sharma2024}, a tendency to persuade rather than be correct \citep{wen2025}, position and verbosity biases \citep{zheng2023}, and susceptibility to superficial qualities \citep{zeng2024}. We mitigate with a \emph{cross-family} judge (a non-Claude model grading a Claude system), blinding to condition, randomized presentation, held-out gold references, and a human inter-rater (Cohen's $\kappa$) audit.

\textbf{Measurement validity for evaluations.} We borrow construct- and threat-to-validity discipline from psychometrics \citep{cronbach1955,messick1995,shadish2002} and the recent LLM-evaluation-validity literature, which finds widespread construct-validity failures in LLM benchmarks \citep{bean2025} and documents benchmark-integrity threats such as contamination \citep{white2025} and reporting-provenance asymmetry \citep{singh2025}. We adopt the discipline directly: define the outcome construct explicitly, enumerate threats, and control them by design (content parity, blinding, preregistration). The study is, in part, an applied case study of this discipline.

\section{System Under Study}

The corpus under study\corpusnote{} is a personal knowledge base of 709 markdown pages (sources, entities, concepts, analyses, proposals, policies) that is continuously maintained by an LLM under a fixed schema. Pages carry YAML frontmatter, an opening lead paragraph, and a \texttt{\#\# Connections} section of inter-page links. It is an instance of the LLM-compiled-wiki architecture that prior work benchmarked against Vector RAG \citep{cochran2026}; here we take that architecture as fixed and study how its internal access structure affects an agent's answers.

The \textbf{progressive-disclosure retrofit} under test comprises three structural changes, layered as ablation arms:
\begin{itemize}
\item \textbf{State extraction (A1):} the volatile ``recent activity'' state that previously lived in the index's frontmatter is moved to a separate \texttt{recent.md}, slimming the index.
\item \textbf{Per-page summaries (A2):} a one-line \texttt{summary:} field is added to every page's frontmatter.
\item \textbf{Retrieval tool (A3):} \texttt{wiki\_query.py}, a keyword-ranked retrieval tool returning ranked page gists (title, summary, and optionally lead paragraph and neighbor links), plus a manifest.
\end{itemize}

The baseline \textbf{A0} is the pre-retrofit corpus. Its routing artifact is the full index: an agent \emph{may} consult it to route and then open pages, but as the pilot showed (\S\ref{sec:intro}), a capable self-routing agent often skips it and reads inferred page paths directly. A0 is therefore a routing-artifact baseline, not a regime in which the agent is guaranteed to load the index.

\section{Method}
\label{sec:method}

\begin{figure}[t]
\centering
\includegraphics[width=0.86\textwidth]{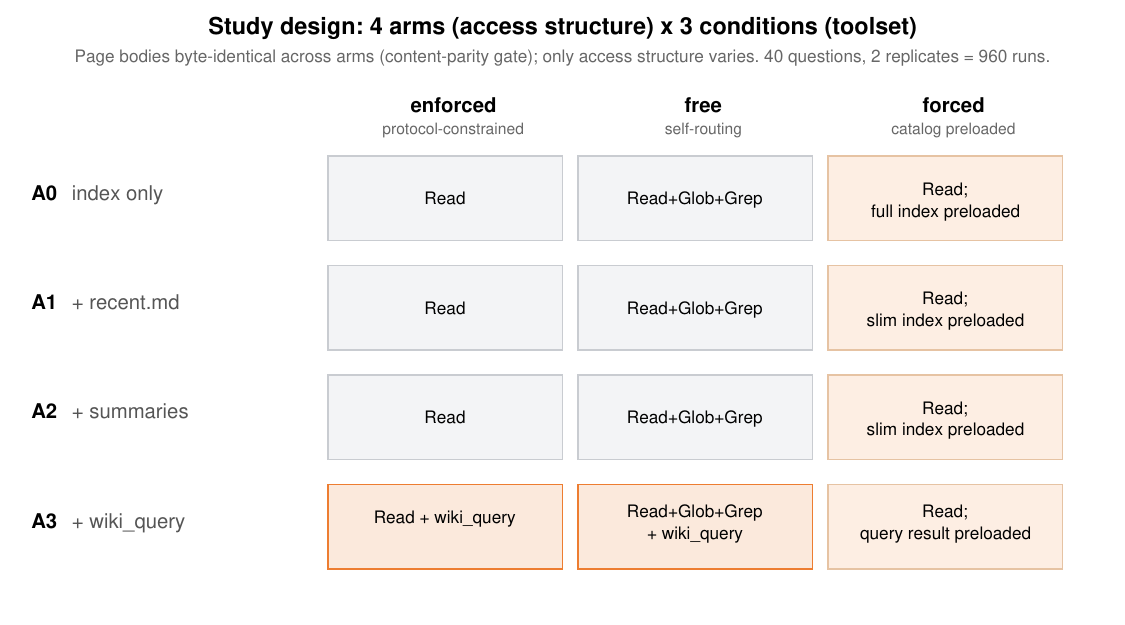}
\caption{Study design. Four arms (access structure) crossed with three tool conditions; page bodies byte-identical across arms.}
\label{fig:design}
\end{figure}

The method is the preregistered plan (OSF \texttt{feka7}, tag \texttt{exp/prereg-v1}); we summarize it here. Figure~\ref{fig:design} shows the arm $\times$ condition design.

\textbf{Design.} A 4 (arm) $\times$ 3 (tool condition) fully-crossed, within-item factorial with 2 replicates per cell. Arms A0--A3 are as above. Conditions: \textbf{enforced} (A0--A2 read-only; A3 may run the retrieval tool: a documented-protocol agent), \textbf{free} (adds general filename/content search: a realistic self-routing agent), and \textbf{forced} (read-only, with the arm's routing artifact pre-loaded into the prompt: A0 receives the original $\sim$150~KB monolithic index, A1 and A2 their correspondingly slimmed $\sim$132~KB index, and A3 only the keyword-retrieval result, a naive catalog-preload pattern with no similarity retrieval). The experimental unit is the question; 40 questions are stratified into four retrieval-breadth types (single-hop, multi-hop, historical-state, aggregation; 10 each). Total $= 40 \times 4 \times 3 \times 2 = \textbf{960 runs}$.

\textbf{Corpus freezing and content parity.} Each arm is an immutable annotated git tag (\texttt{exp/corpus-2026-07-03}, \texttt{exp/arm-a1-recent}, \texttt{exp/arm-a2-summaries}, \texttt{exp/arm-a3-retrieval}). A parity gate (\texttt{check\_parity.py}) verifies that all 709 content-page \textbf{bodies} are byte-identical across arms; only access structure (frontmatter, \texttt{recent.md}, the retrieval tool) differs. Arm tooling is further isolated by the tag; e.g., the retrieval tool exists only in the A3 worktree.

\textbf{Questions.} Stratified purposive sample; roughly half seeded from the corpus's real logged query history (ecological validity), the remainder authored to fill strata. The benchmark is frozen and its SHA-256 registered before any confirmatory run.

\textbf{Runs.} An automated harness launches a fresh, independent headless session of a fixed model (\texttt{claude-opus-4-8[1m]}) per (question $\times$ arm $\times$ condition $\times$ replicate) under the condition's allowed-tool whitelist, and records the answer plus per-run usage from the model API. Execution order over the 960 runs is completely randomized under a fixed seed. Each run is memoryless (no cross-run carryover), so no counterbalancing is needed; randomization controls prompt-cache and temporal nuisance.

\textbf{Measures.} \emph{Primary (quality):} a blinded LLM judge scores each answer on four rubric dimensions (correctness, completeness, citation validity, concision), each 0--2, composite 0--8. \emph{Secondary (cost):} we report cost three ways, to keep the result robust to pricing and cache state (Appendix~\ref{app:cost}): \textbf{billable dollars} (provider-billed \texttt{total\_cost\_usd}), \textbf{logical context tokens} (all tokens supplied to or produced by the model), and a cache-robust \textbf{new-token burden} (uncached input $+$ cache-creation $+$ output, excluding cache-read; this isolates the tokens billed above the cache-read rate). Also recorded per run: tool turns, wall-clock, pages cited, permission denials.

\textbf{Cost model.} Dollar costs are provider-billed \texttt{total\_cost\_usd} for \texttt{claude-opus-4-8[1m]}, not a hardcoded rate; they are the actual billed cost under the price list on the run dates. Because cache state can swing dollars $\sim$12.5$\times$ for identical work, we report the cache-robust new-token burden alongside dollars and, in Appendix~\ref{app:cost}, a cold/warm/actual cache-state sensitivity check that confirms the arm ordering under every cache assumption. All cost figures and Figure~\ref{fig:cost} are answer-generation cost only; the gpt-5 judge's grading cost is separate and excluded.

\textbf{Gold and grading.} For each question, a gold reference answer with a must-include checklist was verified against the frozen corpus and its bytes hashed before grading. The confirmatory judge is a \textbf{cross-family} model (OpenAI gpt-5 \citep{openai2025}, grading a Claude system), blinded to arm/condition, with randomized presentation. A human rater grades a stratified $\ge$15\% subsample; inter-rater agreement is reported as Cohen's $\kappa$, and if $\kappa < 0.60$ the rubric/judge is revised and all answers re-graded (this was the registered rule; the disclosed deviation from it is reported in \S\ref{sec:repro}).

\textbf{Analysis.} \emph{Primary (quality):} a linear mixed-effects model, \texttt{composite \textasciitilde{} arm + condition + (1 \textbar{} question)}; non-inferiority of A3 vs.\ A0 declared if the upper bound of the 95\% CI for (A0 $-$ A3) is below $\delta = 0.5$. The margin $\delta = 0.5$ is 6.25\% of the 0--8 composite scale, equivalently a quarter-point on two of the four rubric dimensions; we fixed it, before the confirmatory runs, as the largest quality loss we would accept in exchange for the cost reductions under test, following non-inferiority reporting guidance that the margin be justified as a substantively acceptable difference rather than chosen post hoc \citep{piaggio2012}. As a sensitivity check we also refit with an arm $\times$ condition interaction and report the A3$-$A0 quality contrast per condition (\S\ref{sec:reliability}). \emph{Secondary (cost):} \texttt{new\_tokens \textasciitilde{} arm * condition + (1 \textbar{} question)} and dollars analogously, reporting the A3$-$A0 effect per condition. Ablation contrasts A0$\to$A1$\to$A2$\to$A3 within each condition. $\delta$, replicate count (2), and the cost DV were fixed by the pilot; the pilot data are excluded from the confirmatory analysis.

\section{Results}

All numbers are from the 960-run confirmatory study graded by the preregistered gpt-5 judge; bootstrap CIs use $B = 5000$ over per-question paired contrasts (seed 20260703).

\subsection{Primary: answer quality (non-inferiority)}

\begin{table}[ht]
\centering
\caption{Mean quality composite (0--8) by arm, pooled across conditions, with the A3$-$A0 contrast and 95\% CI.}
\label{tab:quality}
\begin{tabular}{lcc}
\toprule
Arm & Mean composite & $\Delta$ vs.\ A0 (95\% CI) \\
\midrule
A0 & 6.30 & n/a \\
A1 & 6.35 & $+0.05$ ($-0.16$, $+0.23$) \\
A2 & 6.29 & $-0.01$ ($-0.23$, $+0.20$) \\
A3 & 6.32 & $+0.01$ ($-0.27$, $+0.26$) \\
\bottomrule
\end{tabular}
\end{table}

\textbf{A3 is non-inferior to A0} at $\delta = 0.5$: the A3$-$A0 contrast is $+0.01$ (95\% CI $-0.27$, $+0.26$), so the upper bound of the (A0 $-$ A3) contrast is $+0.27$, below the 0.5 margin. The retrofit is non-inferior but not superior. This pooled result, however, masks a condition-dependent interaction and a small correctness/completeness deficit that the composite offsets; both are reported with sensitivity analyses in \S\ref{sec:reliability}. Per dimension (each 0--2), the A3$-$A0 differences are correctness $-0.09$, completeness $-0.08$, citation validity $+0.05$, and padding $+0.14$ (higher $=$ less padding). A3 thus trades a small amount of correctness and completeness for better citation validity and materially less padding, consistent with the hypothesized mechanism that gist-ranked retrieval and a slug-validating tool produce tighter, better-cited answers. \emph{(The pilot's directional superiority (A3 7.11 vs.\ A0 6.56) did not replicate under the cross-family gpt-5 judge; absolute levels are not comparable across judges, but the within-study arm comparison shows flat quality.)}

\subsection{Secondary: cost by condition}
\label{sec:cost}

\begin{table}[ht]
\centering
\caption{A3-vs-A0 cost effect within each condition, with 95\% CIs.}
\label{tab:cost}
\begin{tabular}{lrrll}
\toprule
Condition & A0 \$ & A3 \$ & A3$-$A0 \$ (95\% CI) & A3$-$A0 new-tok (95\% CI) \\
\midrule
enforced & 0.711 & 0.498 & $-0.213$ ($-0.265$, $-0.160$) & $-21{,}147$ ($-24{,}801$, $-17{,}448$) \\
free     & 0.764 & 0.501 & $-0.263$ ($-0.301$, $-0.227$) & $-26{,}091$ ($-28{,}407$, $-23{,}837$) \\
forced   & 1.108 & 0.468 & $-0.641$ ($-0.700$, $-0.577$) & $-52{,}377$ ($-57{,}822$, $-46{,}690$) \\
\bottomrule
\end{tabular}
\end{table}

\begin{figure}[t]
\centering
\includegraphics[width=0.86\textwidth]{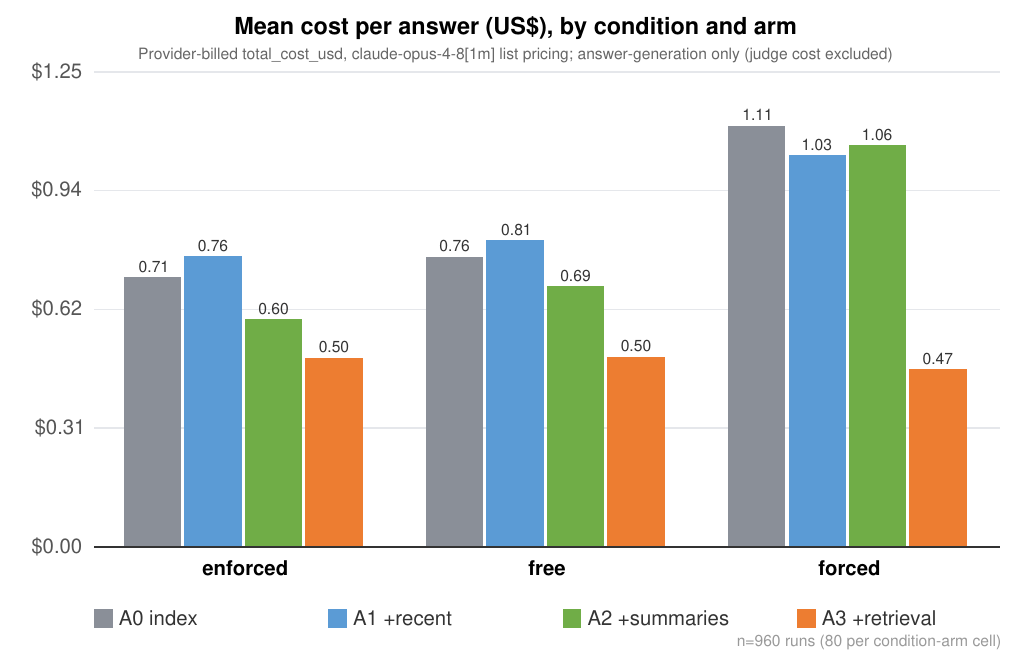}
\caption{Mean answer-generation cost per answer (US\$) by condition and arm, as provider-billed \texttt{total\_cost\_usd} for \texttt{claude-opus-4-8[1m]} at list pricing on the run dates (uncached input \$5, output \$25, cache-write \$6.25, cache-read \$0.50 per MTok; no long-context premium); judge cost excluded. Dollars are cache-state-dependent; see the cache-robust new-token contrasts in Table~\ref{tab:cost} and the cold/warm/actual check in Appendix~\ref{app:cost}. $n = 960$ runs (80 per condition-arm cell).}
\label{fig:cost}
\end{figure}

Contrary to the registered expectation (a null under enforced/free and a saving only under forced), \textbf{A3 is significantly cheaper than A0 in all three conditions}, every dollar and new-token CI excluding zero. The saving is largest under forced ($-\$0.641$, $-58\%$), where A0 preloads the full 150~KB index, but it is real and significant even under enforced ($-\$0.213$, $-30\%$) and free ($-\$0.263$, $-34\%$), where no arm preloads anything. We flag this as a \textbf{deviation from the registered secondary hypothesis, in the favorable direction}: the retrofit reduces cost more broadly than predicted. Because the cache-robust new-token contrasts track the dollar contrasts in sign and significance, the effect is not an artifact of cache-state pricing; a cold/warm/actual sensitivity check (Appendix~\ref{app:cost}) confirms A3 is cheaper than A0 under every cache assumption. Figure~\ref{fig:cost} shows mean cost per answer by condition and arm.

\subsection{Ablation (A0$\to$A1$\to$A2$\to$A3)}

\begin{table}[ht]
\centering
\caption{Mean US\$ by arm within condition.}
\label{tab:ablation}
\begin{tabular}{lcccc}
\toprule
Condition & A0 & A1 & A2 & A3 \\
\midrule
enforced & 0.711 & 0.765 & 0.599 & 0.498 \\
free     & 0.764 & 0.807 & 0.686 & 0.501 \\
forced   & 1.108 & 1.030 & 1.058 & 0.468 \\
\bottomrule
\end{tabular}
\end{table}

The ablation localizes the saving to summaries and the retrieval tool, not state extraction. Moving volatile state out of the index (A1) does not reduce cost and slightly raises it under enforced and free (0.765 vs.\ 0.711; 0.807 vs.\ 0.764); per-page summaries (A2) begin to help (0.599, 0.686); the retrieval tool (A3) helps most in every condition. Under forced, A1 and A2 remain close to A0 because they still preload a monolithic catalog, even though that catalog is slimmed by $\sim$18~KB (132~KB vs.\ A0's 150~KB, enough to shave them slightly below A0: 1.030 and 1.058 vs.\ 1.108); only A3 replaces catalog preload with a per-query retrieval result, collapsing cost to 0.468.

\subsection{Per-stratum and exploratory}
\label{sec:perstratum}

A3 is cheaper than A0 in all four question strata (single-hop \$0.350 vs.\ \$0.747, multi-hop \$0.564 vs.\ \$0.925, historical-state \$0.446 vs.\ \$0.793, aggregation \$0.594 vs.\ \$0.980). Quality is flat or slightly higher for A3 on single-hop (7.00 vs.\ 6.83), multi-hop (6.85 vs.\ 6.62), and historical-state (6.05 vs.\ 6.07), and lower only on aggregation (5.37 vs.\ 5.70), the broad-recall stratum where a narrower cited evidence set costs completeness. The \textbf{cost mechanism} is consistent with more targeted access rather than avoided index-loading. Pages cited per answer falls monotonically from 6.10 (A0) to 6.00 (A1), 5.67 (A2), and 4.22 (A3), a 31\% reduction that mirrors the cost decline; tool turns per answer fall in the same direction (A0 4.98 $\to$ A3 4.45, and within every condition), an independent behavioral signal. We did not capture per-read telemetry (unique pages opened, read-call counts, or read-token totals) because the harness recorded only aggregate usage; pages cited and tool turns are therefore consistent with, but do not by themselves prove, a reduction in pages actually read. Direct read telemetry is a target for replication (\S\ref{sec:limitations}). Self-routing pressure is also visible: even in the read-only enforced condition, agents attempt corpus-wide \texttt{grep}/\texttt{find} searches on 39 of 320 runs (vs.\ 8 free, 3 forced), all blocked by the whitelist (\S\ref{sec:threats}).

\subsection{Grading reliability and sensitivity analyses}
\label{sec:reliability}

The author served as the human rater, independently scoring the preregistered 15\% stratified subsample (144 of 960 runs), blinded to arm and condition, on the same four 0--2 dimensions. A single, potentially non-independent rater is a limitation (\S\ref{sec:limitations}); blinding to arm and condition mitigates but does not remove it. Table~\ref{tab:kappa} reports agreement with the gpt-5 judge.

\begin{table}[ht]
\centering
\caption{Human vs.\ gpt-5 agreement on the 144-run audit subsample.}
\label{tab:kappa}
\begin{tabular}{lcccc}
\toprule
Dimension & Cohen's $\kappa$ (weighted) & Gwet's AC1 & \% exact & \% within 1 \\
\midrule
correctness  & 0.30 & 0.83 & 85\% & 99\% \\
completeness & 0.56 & 0.77 & 81\% & 100\% \\
citation     & 0.10 & 0.44 & 56\% & 96\% \\
padding      & 0.13 & 0.32 & 49\% & 92\% \\
\midrule
\textbf{pooled} & \textbf{0.23} & \textbf{0.60} & \textbf{68\%} & \textbf{97\%} \\
\bottomrule
\end{tabular}
\end{table}

\textbf{The preregistered reliability gate was not met.} The pooled quadratic-weighted Cohen's $\kappa$ is 0.23, far below the planned 0.60 threshold. We therefore report the gpt-5 judge analysis as the \textbf{preregistered first-pass result} and treat the failed reliability audit as a material limitation, not as a passed check. Two mechanisms explain where the shortfall comes from. First, $\kappa$ is prevalence-sensitive: correctness and completeness sit near the ceiling of the scale (85\% and 81\% exact agreement), where $\kappa$ collapses despite high raw agreement (the high-prevalence paradox \citep{feinstein1990}). Correctness and completeness therefore show high raw agreement and high AC1 \citep{gwet2008} (0.83 and 0.77) but not high $\kappa$ (0.30 and 0.56). Second, the genuine disagreement is localized to citation (AC1 0.44) and padding (AC1 0.32), where the author rated more leniently than the judge (mean difference $+0.46$ and $+0.57$): the judge penalizes wiki-style thoroughness as verbosity and requires a citation to directly support its adjacent claim, whereas the human counts responsive thoroughness and resolving links as sufficient. Across all dimensions the raters differ by more than one point on only 3.5\% of scores.

Because the gate failed, we do not rest the quality conclusion on the pooled composite alone. Table~\ref{tab:sensitivity} reports three sensitivity analyses of the A3$-$A0 quality contrast.

\begin{table}[ht]
\centering
\caption{Sensitivity analyses for the A3$-$A0 quality contrast (bootstrap 95\% CIs, $B = 5000$).}
\label{tab:sensitivity}
\begin{tabular}{@{}p{5.3cm}ccp{5.0cm}@{}}
\toprule
Analysis & A3$-$A0 & 95\% CI & Non-inferiority at $\delta = 0.5$ \\
\midrule
Full composite, judge, all 960 & $+0.01$ & ($-0.26$, $+0.26$) & holds \\
Correctness$+$completeness only, judge, all 960 & $-0.18$ & ($-0.33$, $-0.05$) & holds at $\delta{=}0.5$; fails at scaled $\delta{=}0.25$ \\
Full composite, human subsample (144) & $+0.05$ & ($-0.34$, $+0.40$) & holds (human rates A3 marginally higher) \\
Corr$+$compl, human subsample (144) & $-0.22$ & ($-0.58$, $+0.09$) & not established; CI spans no effect and a loss beyond $\delta$ \\
Full composite (enforced only) & $+0.13$ & ($-0.20$, $+0.41$) & holds \\
Full composite (free only) & $+0.30$ & ($+0.03$, $+0.58$) & holds (A3 significantly better) \\
Full composite (forced only) & $-0.39$ & ($-1.01$, $+0.19$) & not established (point favors A0; CI includes 0) \\
\bottomrule
\end{tabular}
\end{table}

Two robust facts and one genuine caveat emerge. (i)~Full-composite non-inferiority holds under \textbf{both} graders independently, and the human, if anything, rates A3 higher. (ii)~It also holds under the two realistic self-routing regimes (enforced, free), where A3 is non-inferior or significantly better. (iii)~But restricting to the two AC1-reliable dimensions, correctness$+$completeness, exposes a small, statistically significant A3 deficit (judge $-0.18$, CI excludes zero) that the full composite offsets with A3's citation and padding gains; and non-inferiority is not established under \textbf{forced} catalog-preload ($-0.39$, CI $-1.01$ to $+0.19$, which includes zero, so the point estimate favors A0 without demonstrating inferiority), where A0--A2 receive the whole catalog and A3 only the retrieval hits. The non-inferiority conclusion is thus margin- and condition-dependent on the reliable dimensions: solid at $\delta = 0.5$ overall and under self-routing access, but not established under forced preload or at a stricter scale-proportional margin.

We treat this as a disclosed deviation from the registered reliability rule (\S\ref{sec:repro}). The sensitivity analyses separate the robust parts of the quality conclusion from the dimensions and conditions where it is margin-sensitive. Because the judge was blind to arm, a re-grade with tightened anchors would most directly affect absolute citation and padding levels; it would not eliminate the observed correctness/completeness sensitivity, which we therefore report separately. The two anchors that drove the $\kappa$ shortfall do need tighter operational definitions: for a 2 on citation, that the cited page directly establish the specific claim; for a 2 on padding, that thoroughness be responsive to the question, with length alone never the criterion. We flag those for replication (\S\ref{sec:limitations}).

\section{Discussion}

\textbf{Measured versus nominal.} The saving the retrofit was built to capture, avoiding a 150~KB index load on every query, is real only in the catalog-preload (forced) regime, where it drives a 58\% cost drop. A capable self-routing agent never loads the index, so that specific saving does not materialize for it, exactly as the pilot found. Yet the confirmatory data show the retrofit still cuts total cost in every regime, 30\% under enforced and 34\% under free, through a different mechanism: gist-ranked retrieval and per-page summaries are associated with more targeted access, with the agent citing fewer pages (6.10 down to 4.22) and taking fewer tool turns, at non-inferior quality under those regimes. The nominal index-load saving is thus regime-specific rather than deployment-general; the benefit that actually shows up in deployment is more targeted access, real for a different reason than advertised. The lesson for evaluation validity holds, and is sharper for being quantified: the claimed mechanism did not survive contact with how a capable agent uses the corpus, and only a controlled measurement separated the nominal component from the real one.

\textbf{When does progressive-disclosure structure pay off?} The forced-regime result shows that catalog-preload pipelines, and by extension agents that cannot self-route, get the largest \emph{cost} benefit, but, as \S\ref{sec:reliability} shows, forced is also the one regime where A3 fails the non-inferiority criterion: the point estimate favors A0 ($-0.39$ composite), though the CI includes zero, so this is a failure to establish non-inferiority, not a demonstrated harm. Catalog-preload feeds A0--A2 the whole catalog while A3 sees only its retrieval hits. The forced regime therefore buys the biggest saving where the quality trade is least favorable; the clearly favorable trade lives in the self-routing regimes. There, unlike the pilot-driven prediction of a near-null, even a capable self-routing agent saves roughly a third (30\% enforced, 34\% free) from summaries plus retrieval, at non-inferior or better quality. The capability-dependence prediction therefore sharpens rather than disappears: weaker agents should benefit more because they cannot route around a monolithic index, but the floor is not zero. A preregistered two-model design (capable vs.\ weaker agent) remains the natural follow-up.

\textbf{Quality.} Leaner, gist-ranked access does not degrade overall quality (non-inferior at $\delta = 0.5$) and shifts the dimension profile rather than lowering it: better citation validity ($+0.05$) and less padding ($+0.14$), at a small correctness ($-0.09$) and completeness ($-0.08$) cost concentrated in broad-recall aggregation questions. On the two most reliably-graded dimensions alone that small correctness/completeness cost is statistically detectable (\S\ref{sec:reliability}); the composite is flat because A3's citation and concision gains offset it. The citation-validity gain matches the retrofit's design intent, since per-page summaries and a slug-validating retrieval tool make it easier to cite pages that exist and support the claim; the aggregation-completeness dip is the cost of narrower evidence use when a question needs broad coverage.

\section{Threats to Validity}
\label{sec:threats}

\begin{itemize}
\item \textbf{Content-parity confound (central):} controlled by construction, with byte-identical bodies across arms, gated before analysis.
\item \textbf{Judge bias:} cross-family judge, blinding, randomized presentation, held-out hashed gold, human $\kappa$ audit.
\item \textbf{Contamination/carryover:} fresh, memoryless sessions per run; no cross-run state.
\item \textbf{Cache/order nuisance:} randomized execution order under a fixed seed; cost reported in dollars (paid) alongside the cache-robust new-token burden, with a cold/warm/actual cache-state sensitivity check (Appendix~\ref{app:cost}).
\item \textbf{Construct validity of ``quality'':} a four-dimension rubric with per-dimension reporting, audited against a human rater; the preregistered reliability gate failed, and the quality conclusion is supported by the sensitivity analyses in \S\ref{sec:reliability} rather than a passed audit.
\item \textbf{Tool-integrity:} per-condition allowed-tool whitelists are enforced at the tool-call site, so a disallowed call is blocked before it executes and recorded as a \texttt{permission\_denial} (a preregistered secondary measure); no disallowed tool runs, and no run is discarded on this basis. The denial rate is itself reported as evidence of self-routing pressure: even under the read-only \emph{enforced} condition, agents attempt corpus-wide \texttt{grep}/\texttt{find} searches (39 of 320 enforced runs, vs.\ 8 free and 3 forced), and the whitelist blocks them.
\end{itemize}

\section{Limitations and Future Work}
\label{sec:limitations}

Single corpus, single answering model, one author's question distribution, and a single non-independent human rater (the author): causal claims are about access structure \emph{within this system}; external validity to other corpora, models, users, and raters is not established. The design deliberately trades breadth for internal validity. The harness also recorded only aggregate token usage, not per-tool-call read telemetry, so the mechanism evidence rests on pages cited and tool turns rather than direct read counts (\S\ref{sec:perstratum}). Natural extensions: a preregistered \textbf{two-model} design (capable vs.\ weaker agent) to test the predicted capability-dependence of the cost effect; multiple corpora; independent human raters; end-users in place of an automated agent; per-read telemetry to test the narrower-evidence-use mechanism directly; and a grading rubric with tightened citation and concision anchors (\S\ref{sec:reliability}) to improve human-judge convergence on those two dimensions.

\textbf{Where the result should be expected to travel.} None of the following is tested here, and we state them as reasoned expectations rather than findings. \emph{Agent capability} is the moderator the design already probes: the forced condition stands in for an agent that cannot self-route, and it is simultaneously where the cost saving is largest and where the quality trade is least favorable, so we expect the cost effect to grow and the quality margin to tighten as answering agents get weaker. \emph{Corpus size} should move the two cost components differently. The nominal index-load saving scales with catalog size, so it is larger above our 709 pages and vanishes for corpora small enough that the index costs little to load, a threshold we did not attempt to locate. The targeted-access saving instead depends on how well one-line gists discriminate among pages, which is a property of summary quality rather than of page count. \emph{Domain} is where we can say least: this corpus is prose-dense and densely cross-linked, and a corpus of short, weakly-linked, or highly templated records would exercise both mechanisms differently enough that we would not extrapolate to it. The two-model design is the cheapest of the three to test and the one whose answer we would most expect to change deployment decisions.

\section{Reproducibility and Data Availability}
\label{sec:repro}

Preregistration: \osflink, registered public 2026-07-03; the plan was locked the same day at git tag \texttt{exp/prereg-v1}. OSF registrations are frozen at registration time and cannot accept later uploads, and the plan frozen on this registration is the pre-lock draft, uploaded from a stale working copy after the lock. The draft carries the same hypotheses, analysis model, provisional margin $\delta = 0.5$, cross-family judge and stratified human audit. It differs in being marked as a draft, in specifying three replicates per cell (1,440 runs) where the locked plan fixes two (960 runs), and in omitting the pilot section (\S5a of the locked plan) that records the power calculation behind that change. The locked plan and the materials current at submission are deposited on the parent project, \materialslink, and the manifest in \texttt{MATERIALS.md} enumerates them. The corpus is frozen at tags \texttt{exp/corpus-2026-07-03} (A0), \texttt{exp/arm-a1-recent}, \texttt{exp/arm-a2-summaries}, \texttt{exp/arm-a3-retrieval} in a private repository (the corpus contains internal material). The harness, analyzer, grading-judge runner, parity gate, rubric, human-audit tool, per-cell baselines, operational runbook, and per-arm/condition protocols are released. Held-out grading materials (40 gold answers $+$ the question set) are withheld to protect internal content and preserve blind grading; their SHA-256 manifest is registered (combined digest \texttt{5573ed6f2bc05508df8ce3092a723369\allowbreak fdd24682f75c539c9e7e3f71f2665e46}) and released with the paper, so readers can verify no post-hoc edits. The per-run agent answers and judge grades are likewise withheld, as they contain corpus-derived text; results are reported in aggregate. The answering model is pinned: \texttt{claude-opus-4-8} served at the 1M-context tier (recorded by the harness as \texttt{claude-opus-4-8[1m]}); the confirmatory judge is the OpenAI \texttt{gpt-5} endpoint; the confirmatory runs and grading were executed on 2026-07-03/04; and the randomization seed (20260703) is fixed. The released harness (\texttt{run\_harness.py}, \texttt{grade.py}) records these exact model strings, tool whitelists, and run parameters for both the answering model and the judge.

\textbf{Deviations from the preregistration.} Three; the first two are also disclosed at the point of use. (1)~The registered secondary cost hypothesis predicted a null cost difference under the enforced and free conditions and a saving only under forced; the confirmatory data refute this in the favorable direction, with A3 significantly cheaper in all three conditions (\S\ref{sec:cost} and Table~\ref{tab:cost}). (2)~The registered reliability rule specified that a human-judge Cohen's $\kappa$ below 0.60 triggers a rubric revision and full re-grade; the pooled $\kappa$ was 0.23. Rather than re-grade, we report the gpt-5 result as the preregistered first-pass analysis, treat the failed gate as a material limitation, and support the quality conclusion with sensitivity analyses (\S\ref{sec:reliability}, Table~\ref{tab:sensitivity}): paradox-robust AC1, a correctness$+$completeness-only cut, the human-audited subsample, and a per-condition (arm $\times$ condition) breakdown. We take this path because the shortfall is attributable to the $\kappa$ prevalence paradox on two near-ceiling dimensions plus a diagnosable under-specification of two rubric anchors, and because the judge's blindness to arm means a re-grade could not alter the one quality contrast (correctness$+$completeness) that is margin-sensitive. (3)~The registered document specifies three replicates per cell (1,440 runs); the study ran two (960 runs). The reduction was fixed in the locked plan by the pilot power analysis in its \S5a, where a pooled SD of about 1.19 and $\delta = 0.5$ require about 90 observations per arm for 80\% power against the 240 per arm collected. Fewer replicates widen the confidence intervals, which makes the primary non-inferiority test harder to pass, not easier.

\section{Conclusion}

On a real 709-page LLM-maintained knowledge base, restructuring for progressive disclosure leaves answer quality non-inferior overall (A3$-$A0 $= +0.01$ composite, well within $\delta = 0.5$; non-inferior or better under self-routing access; under forced catalog-preload, the point estimate favored A0 and non-inferiority was not established; and the preregistered reliability gate failed, so the quality conclusion rests on the \S\ref{sec:reliability} sensitivity analyses rather than on a passed audit) while cutting per-question cost in every access regime: 30\% under enforced, 34\% under free, and 58\% under forced. The saving does not come from the originally-claimed avoidance of a monolithic index load, which is real only in the catalog-preload regime. It is instead associated with more targeted access: the agent cites fewer pages (6.10 to 4.22) and takes fewer tool turns, via per-page summaries and a keyword-ranked retrieval tool. The registered expectation of a cost-null for self-routing agents was refuted in the favorable direction. The methodological lesson is the measured-versus-nominal gap: an efficiency claim whose stated mechanism did not survive contact with how a capable agent uses the corpus, even though the benefit itself did, is exactly the kind of gap a preregistered, content-parity-controlled measurement is built to expose.

\bibliography{\bibpath}

\appendix

\section{Cost model, metrics, and cache-state sensitivity}
\label{app:cost}

We do not compute dollar cost from a hardcoded token rate. The dollar figure for each run is the provider-billed \texttt{total\_cost\_usd} reported by the model API/CLI for \texttt{claude-opus-4-8[1m]}, which applies the provider's per-token schedule to that run's exact token mix. Reported dollars are therefore the actual billed cost under the price list in effect on the run dates (2026-07-03/04). Those list rates, per million tokens: uncached input \$5.00, output \$25.00, cache write (5-minute TTL) \$6.25 (1.25$\times$ input), cache read \$0.50 (0.10$\times$ input) \citep{anthropic-pricing,anthropic-caching}. The \texttt{[1m]} suffix denotes the 1M-context tier, which for Opus 4.8 serves at standard pricing (no long-context surcharge), so the rates hold across our context sizes.

We report cost on three complementary metrics:
\begin{itemize}
\item \textbf{Billable dollars:} provider-reported \texttt{total\_cost\_usd}; the paid cost, and cache-state-dependent.
\item \textbf{Logical context tokens:} uncached input $+$ cache-creation $+$ cache-read $+$ output; every token supplied to or produced by the model, regardless of billing rate.
\item \textbf{New-token burden:} uncached input $+$ cache-creation $+$ output, excluding cache-read. This isolates the tokens billed above the cache-read rate. It is \emph{not} ``physical work'' (cache-read tokens are still real context), but it is a cache-state-robust quantity reconstructable under any price list, and it is the pricing-independent DV reported alongside dollars.
\end{itemize}

Because a cached token bills at 0.10$\times$ when read but 1.25$\times$ when first written, identical work can differ $\sim$12.5$\times$ in dollars depending on cache state (cold vs.\ warm), a function of run timing, not of the arm. To show the cost result is not a cache-pricing artifact, Table~\ref{tab:cache} recomputes mean cost per answer under three standardized cache assumptions: \textbf{all-cold} (every input token billed at the \$5 base rate), \textbf{all-warm} (every input token billed at the \$0.50 cache-read rate), and \textbf{actual billed}. A3 is cheaper than A0 under all three.

\begin{table}[ht]
\centering
\caption{Mean answer-generation cost per answer (US\$), pooled across conditions, under three cache assumptions.}
\label{tab:cache}
\begin{tabular}{lccc}
\toprule
Arm & all-cold & all-warm & actual billed \\
\midrule
A0 & 1.401 & 0.191 & 0.861 \\
A1 & 1.377 & 0.190 & 0.867 \\
A2 & 1.184 & 0.170 & 0.781 \\
A3 & 0.884 & 0.134 & 0.489 \\
\bottomrule
\end{tabular}
\end{table}

A3 vs.\ A0: 37\% cheaper under all-cold, 30\% under all-warm, 43\% under actual billing. The arm ordering (A3 $<$ A2 $<$ A0 $\approx$ A1) is preserved under every cache assumption.

This new-token definition also resolves a cost-telemetry ambiguity flagged as a limitation in prior work on this architecture \citep{cochran2026}, where uncached, cache-creation, and cache-read tokens were aggregated into a single field despite cache-reads billing at roughly a tenth of the base rate. All cost figures and Figure~\ref{fig:cost} are answer-generation cost only; the cross-family gpt-5 judge's grading cost is separate and excluded.

\end{document}